\documentclass[10pt,twocolumn,letterpaper]{article}

\usepackage{iccv}
\usepackage{times}
\usepackage{epsfig}
\usepackage{graphicx}
\usepackage{amsmath}
\usepackage{amssymb}

\usepackage{mathtools}
\usepackage{bbold}
\usepackage{algorithm}
\usepackage{algpseudocode}
\usepackage{multirow}
\usepackage{enumitem}
\usepackage{color}

\newcommand{\doublecheck}[1]{\textcolor{black}{#1}}
\newcommand{\keypoint}[1]{\vspace{0.1cm}\noindent\textbf{#1}\quad}
\newcommand{\cut}[1]{}
\usepackage{textcomp}
\usepackage{bm}

\DeclareMathAlphabet\mathbfcal{OMS}{cmsy}{b}{n}
\usepackage[pagebackref=true,breaklinks=true,letterpaper=true,colorlinks,bookmarks=false]{hyperref}

\iccvfinalcopy % *** Uncomment this line for the final submission

\def\iccvPaperID{4391} % *** Enter the ICCV Paper ID here

\ificcvfinal\fi

\begin{document}

%%%%%%%%% TITLE
\title{Text is Text, No Matter What: \\ Unifying Text Recognition using Knowledge Distillation}

 \author{Ayan Kumar Bhunia\textsuperscript{1} \hspace{.2cm}  Aneeshan Sain\textsuperscript{1,2}\hspace{.2cm} Pinaki Nath Chowdhury\textsuperscript{1,2}\hspace{.2cm}   Yi-Zhe Song\textsuperscript{1,2} \\
\textsuperscript{1}SketchX, CVSSP, University of Surrey, United Kingdom. 
\\
\textsuperscript{2}iFlyTek-Surrey Joint Research Centre
on Artificial Intelligence. 
% \textsuperscript{2}University of Edinburgh, United Kingdom.\\
\\{\tt\small \{a.bhunia, p.chowdhury, a.sain, y.song\}@surrey.ac.uk}.  
% \\ {\tt\small\{shuvozit.ghose,  kumar.amandeep015\}@gmail.com.}
}

\maketitle
\ificcvfinal\thispagestyle{empty}\fi

%%%%%%%%% ABSTRACT
\begin{abstract}

Text recognition remains a fundamental and extensively researched topic in computer vision, largely owing to its wide array of commercial applications. The challenging nature of the very problem however dictated a fragmentation of research efforts: Scene Text Recognition (STR) that deals with text in everyday scenes, and Handwriting Text Recognition (HTR) that tackles hand-written text. In this paper, for the first time, we argue for their unification -- we aim for a single model that can compete favourably with two separate state-of-the-art STR and HTR models. We first show that cross-utilisation of STR and HTR models trigger significant performance drops due to differences in their inherent challenges. 
%While unification of these two scenarios is lucrative for commercial applications, this task of multi-scenario text recognition is non-trivial due to inherent domain gap between scene and handwritten texts, language diversity or model capacity limitations.
We then tackle their union by introducing a knowledge distillation (KD) based framework. 
%The network unifies individual STR and HTR models into a single model, resulting in a resource-economic online serving solution. 
This however is non-trivial, largely due to the \textit{variable-length} and sequential nature of text sequences, which renders off-the-shelf KD techniques that mostly works with global fixed length data inadequate. For that, we propose four distillation losses all of which are specifically designed to cope with the aforementioned unique characteristics of text recognition. Empirical evidence suggests that our proposed unified model performs on par with individual models, even surpassing them in certain cases. Ablative studies demonstrate that naive baselines such as a two-stage framework, multi-task and domain adaption/generalisation alternatives do not work as well, further authenticating our design.
%verifying the  appropriateness of our design. 
% at par or slightly outperforms individual models.

\end{abstract}

%%%%%%%%% BODY TEXT

\vspace{-0.5cm}
%%%%%%%%% BODY TEXT
\section{Introduction}
 
Text recognition has been studied extensively in the past two decades ~\cite{long2018scene}, mostly due to its potential in commercial applications. Following the advent of deep learning, great progress \cite{bai2018edit, litman2020scatter, wan2020vocabularyReliance, yu2020towards, bhunia2021unseen, bhunia2021joint, bhunia2021metahtr} has been made in recognition accuracy on different publicly available benchmark datasets \cite{IIIT5K-Words, WangICCV2011, ICDAR2015, IamDataset}. Beyond supervised text recognition,  very recent  attempts have been made that utilise synthetic training data via domain adaptation \cite{zhang2019cvprDAtext}, learn optimal augmentation strategy \cite{luo2020learn, bhunia2019handwriting}, couple with visual question answering \cite{biten2019vqa}, and withhold adversarial attacks \cite{xu2020machines}. 
 
\begin{figure}
\begin{center}
  \includegraphics[width=\linewidth]{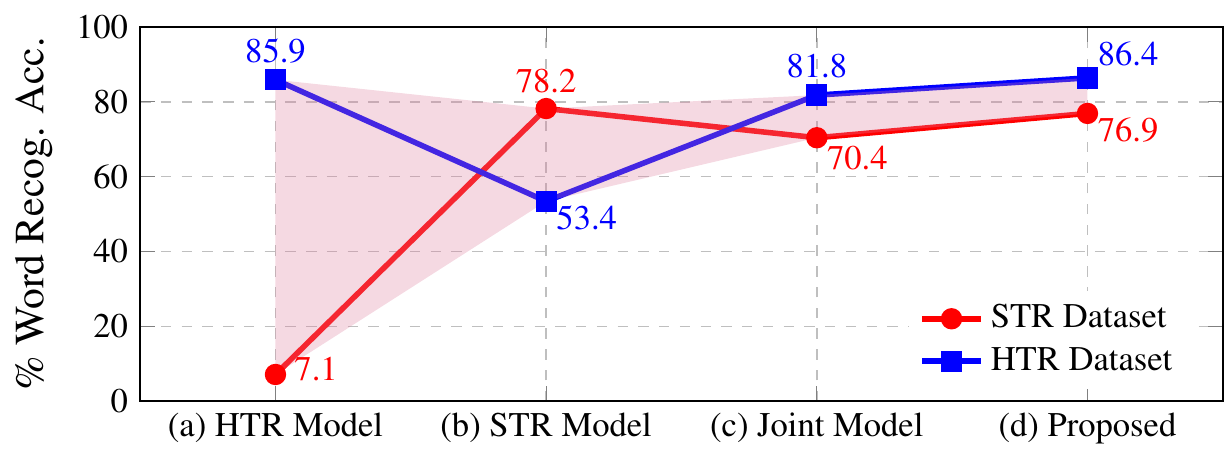} % KD_intro
\end{center}
\vspace{-.15in}
% Take home message
  \caption{Despite performing well for scene images (IAM \cite{IamDataset}), a model trained on HTR datasets (a), performs poorly in STR scenarios (ICDAR-2015 \cite{ICDAR2015}) and vice-versa (b). Although jointly training a model (c) using both STR and HTR datasets helps improve the disparity between the datasets, the gap still remains far behind the specialist models. Our KD based proposed method leads to performance at par or even better than individual models.}
\vspace{-.18in}
\label{fig:Fig1}
\end{figure}

Albeit with great strides made, the field of text recognition remains fragmented, with one side focusing on Scene Text Recognition (STR) \cite{ICDAR2015}, and the other on Handwriting Text Recognition (HTR) \cite{IamDataset}. This however is not surprising given the differences in the inherent challenges found in each respective problem: STR studies text in scene images posing challenges like complex backgrounds, blur, artefacts, uncontrolled illumination \cite{yu2020towards}, whereas HTR tackles handwritten texts where the main challenge lies with the free-flow nature of writing \cite{bhunia2019handwriting} of different individuals. As a result, utilising models trained for STR on HTR (and vice versa) straightforwardly would trigger a significant performance drop (see Figure \ref{fig:Fig1}). This leads to our motivation -- how to design a unified text recognition model that works ubiquitously across both scenarios.

% In this paper, we introduce a new perspective towards text recognition -- unifying multi-scenario text recognition models. 

%a naive approach to train
% Multi-scenario text recognition is deceptively non-trivial due to inherent domain gap between scene and handwritten texts, and variable length sequential nature of word images. 
While there is no existing work addressing this issue, one might naively think of  training a single text recognition network using training data from \emph{both} STR and HTR datasets. However, for the apparent issues of large domain gap and model capacity limitation \cite{tan2019multilingualKD}, while the jointly trained model reduces the performance gap between HTR and STR datasets, it still lags significantly behind individual specialised models. Another solution is to include a classification network prior to specialised STR and HTR models (i.e., a two-stage network). During evaluation, the classifier decides if an input belongs to scene or handwritten text, followed by choosing an appropriate model for downstream recognition. Yet, this solution has two downsides: a) classification network will incur additional computational cost and extra memory consumption to store all three neural networks. b) cascaded connection of the classifier and text recognition models will compound cumulative errors.

In this work, we introduce a \emph{knowledge distillation} (KD) \cite{hinton2015kd, ruffy2019survey} based framework to unify individual STR and HTR models into a \emph{single} multi-scenario model. Our design at a high-level, does not deviate much from a conventional KD setting where a learnable student model tries to mimic the behaviour of a pre-trained teacher. We first train both STR and HTR models separately using their respective training data. Next, each individual model takes turns to act as a teacher in the distillation process, to train a single unified student model. It is this 
{transfer of knowledge} captured by specialised teachers into a single model, that leads to our superior performance in contrast to training a single model using joint STR and HTR datasets (see Figure \ref{fig:Fig1}). 

Making such a design (KD) to work with text recognition is however non-trivial. The difficulty mainly arises from the {variable-length} and sequential natures of text images -- each consists of a sequence of different number of individual characters. Hence, employing off-the-shelf KD methods \cite{ruffy2019survey} that aim at matching output probabilities and/or hidden representations between pre-trained teacher and learnable student model, which are used for global fixed length data, may not be sufficient to transfer knowledge at local character level. We thus propose \emph{three} additional distillation losses to tackle the unique characteristics of text recognition. 
 
More specifically, 
%inspired by attentional decoders that can automatically (weakly-supervised) extract most relevant regions of text image while predicting a particular character in the sequence decoding process. Thus, we obtain localised character-aligned visual feature called glimpse vector, pon which 
we first impose a \emph{character aligned hint loss}. This encourages the student to mimic character-specific hidden representations of specialised teacher over the varying sequence of characters in a text image. Next, an \emph{attention distillation loss} is further imposed over the attention map obtained at every step of character decoding process by an attentional decoder. This compliments the character localised hint-loss, as attention-maps capture rich and diverse contextual information emphasising on localised regions \cite{Hou2019lane}. 
% While the previous two losses focus on localised character level information, capturing the long-range non-local dependencies among the sequential characters is essential, especially for an auto-regressive decoding framework followed by attentional text recognition methods \cite{2dAtten2019aaai}. \blue{<reframe>}
Besides localised character level information, capturing long-range non-local dependencies among the sequential characters is of critical importance, especially for an auto-regressive attentional decoder framework \cite{2dAtten2019aaai}. Accordingly we propose an \emph{affinity distillation loss} as our third loss, to capture the interactions between every pair of positions of the variable character length sequence, and guide the unified student model to emulate the affinity matrix of the specialised teachers. Finally, we also make use of state-of-the-art \emph{logit distillation loss} to work with our three proposed losses. It aims at matching output probabilities of student network over the character vocabulary, with that of pre-trained teachers.

% Using domain specialised models to extract data distribution for supervision of a unified 

Our main contributions can be summarised as follows: (a) We design a practically feasible \emph{unified} text recognition setting that asks a single model to perform equally well across both HTR and STR scenarios. (b) We introduce a novel knowledge distillation paradigm where an unified student model learns from two pre-trained teacher models specialised for STR and HTR. (c) We design three additional distillation losses to specifically tackle the {variable-length} and sequential nature of text data. 
% (b) To realise this objective, we learn an unified student model from two pre-trained teacher models specialised for STR and HTR via a novel knowledge distillation setting for sequential text image.  
(d) Extensive experiments coupled with ablative studies on public datasets, demonstrate the superiority of our framework.
% <basic done> 

\vspace{-0.1cm}
\section{Related Works}

\noindent \textbf{Text Recognition:} With the inception of deep learning, Jaderberg \etal  \cite{jaderberg2014deep, jaderberg2014constrained} introduced a dictionary-based text recognition framework employing deep networks. Alternatively, Poznanski \etal \cite{poznanski2016cnn} addressed the added difficulty in HTR by using a CNN to estimate an n-gram frequency profile.
% which was extended to unconstrained lexicon-free framework \cite{jaderberg2015unconstrained}.
% With the inception of deep learning, Jaderberg \etal  \cite{jaderberg2014deep, jaderberg2014constrained} introduced a text recognition framework employing deep networks, which was restricted to  words only. It was extended to unconstrained lexicon-free framework \cite{jaderberg2015unconstrained}, but was still limited by character level localisation for training. 
Later on, connectionist temporal classification (CTC) layer \cite{GravesICML2006ctc} made end-to-end sequence discriminative learning possible. Subsequently, CTC module was replaced by attention-based decoding mechanism \cite{LeeCVPR2016, shi2017tpami} that encapsulates  language  modeling,  weakly  supervised  character detection  and  character  recognition  under a single model. Needless to say attentional decoder became the state-of-the-art paradigm for text recognition for both scene text \cite{litman2020scatter, yu2020towards, yang2019symmetry, VeriSimilarECCV2018} and handwriting \cite{bhunia2019handwriting, luo2020learn, wang2020decoupled, zhang2019cvprDAtext}. Different incremental propositions \cite{bhunia2021unseen, bhunia2021joint, bhunia2021metahtr} have been made like, improving the rectification module \cite{VeriSimilarECCV2018, yang2019symmetry}, designing multi-directional convolutional feature extractor \cite{AONRecogCVPR2018}, improving attention mechanism \cite{ChengICCV2017, 2dAtten2019aaai} and stacking multiple BLSTM layer for better context modelling \cite{litman2020scatter}.  

Besides improving word recognition accuracy, some works have focused on improving performance in low data regime by designing adversarial feature deformation module \cite{bhunia2019handwriting}, and learning optimal augmentation strategy \cite{luo2020learn}, towards handling adversarial attack \cite{xu2020machines} for text recognition. Zhang \etal \cite{zhang2019cvprDAtext} introduced unsupervised domain adaptation to deal with images from new scenarios, which however definitely demands a fine-tuning step to specialise in new domain incurring additional server costs. On the contrary, we focus on unifying a single model capable of performing consistently well across both HTR and STR images.

\keypoint{Knowledge Distillation:} Earlier, knowledge distillation (KD) was motivated towards training smaller student models from larger teacher models for cost-effective deployment. Caruana and his collaborators \cite{ba2014compression} pioneered in this direction, by using mean square error with the output \emph{logits} of deeper model to train a shallower one. The seminal work by Hinton \etal \cite{hinton2015kd} introduced \emph{softer probability distribution} over classes by a temperature controlled softmax layer for training smaller student models. Furthermore, Romero \etal \cite{romero2015fitnets} employed features learned by the teacher in the intermediate layers, to act as a hint for student's learning. Later works explored different ideas like mimicking \emph{attention maps} \cite{zagoruyko2017KDattention} from powerful teacher, transferring \emph{neuron selectivity} pattern \cite{huang2017selectivity} by minimising Maximum Mean Discrepancy (MMD) metric,  \emph{graminian matrices}  \cite{yim2017gramian} for faster knowledge transfer, multiple \emph{teacher assistants}  \cite{mirzadeh2020TA} for step-wise  knowledge distillation and so on. 
\cut{A radically different approach was taken up in \emph{relational knowledge} distillation \cite{park2019rkd}, which explored structural relationship between teacher and student model via distance and angle loss, instead of focusing on individual outputs.}
% A deeper dive by \cite{heo2019AB} reveals margin ReLU activation, specific position with partial $L_2$ distance for optimised \emph{feature distillation} process. 
In addition to classification setup, KD has been used in object detection \cite{deng2019video}, semantic segmentation \cite{he2019semantic}, depth-estimation \cite{pilzer2019depth}, pose estimation \cite{nie201zeng2019wsod9pose}, lane detection \cite{Hou2019lane}, neural machine translation \cite{tan2019multilingualKD} and so forth.  Vongkulbhisal \etal \cite{vongkulbhisal2019heterogeneous} proposed a methodology of \emph{unifying heterogeneous classifiers} having different label set, into a single unified classifier. 
%  using a generic set of unlabelled data
In addition to obtaining smaller fast-to-execute model, using KD in \emph{self-distillation} \cite{bagherinezhad2018refinery} improves performance of student having identical architecture like teacher. Keeping with self-distillation \cite{bagherinezhad2018refinery}, our teacher networks and trainable student share exactly same architecture, but our motivation lies towards obtaining an unified student model  from  two pre-trained specialised teachers.  
% Here the classifiers have different network architectures and are trained for different target labels.

\keypoint{Unifying models:} A unified model bestows several benefits compared to specialised individual models such as lower annotation and deployment cost as unlike it's counterpart, unified models need not grow linearly with increasing domains \cite{rebuffi2017residualAdapters} or tasks \cite{zamir2018taskonomy} while simultaneously cherishing the benefits of shared supervision. Towards embracing the philosophy of general AI, where the goal is to develop a single model handling multiple purposes, attempts have been made towards solving multiple tasks \cite{kaiser2017oneModel, kokkinos2017ubernet, zamir2018taskonomy} via \emph{multi-task learning}, working over multiple domains \cite{bilen2017missingLink, rebuffi2017residualAdapters}, and employing \emph{universal adversarial attack} \cite{liu2019UnivPerturbPrior}. While unsupervised \emph{domain adaptation} \cite{tzeng2017adversarial} still needs fine-tuning over target domain images, \emph{domain generalisation} \cite{dou2019domain} aims to extract domain invariant features, eliminating the need of post-updating step. \cut{Keeping with our idea, unified VQA mode \cite{shrestha2019answerAll} aim at understanding both natural images and complex synthetic datasets, which require complex chains of compositional reasoning.} In NLP community,   handling multiple language pairs in one model via multi-lingual neural-machine-translation \cite{gu2018unmtLowResource, tan2019multilingualKD}, has been a popular research direction in the last few years. Albeit all these text recognition and \emph{model unifying} approaches are extensively studied topics, we introduce an entirely new aspect of text recognition by unifying STR and HTR scenarios into a single model having significant commercial advantage.

\begin{figure*}[t]
\begin{center}
    \includegraphics[width=0.95\linewidth]{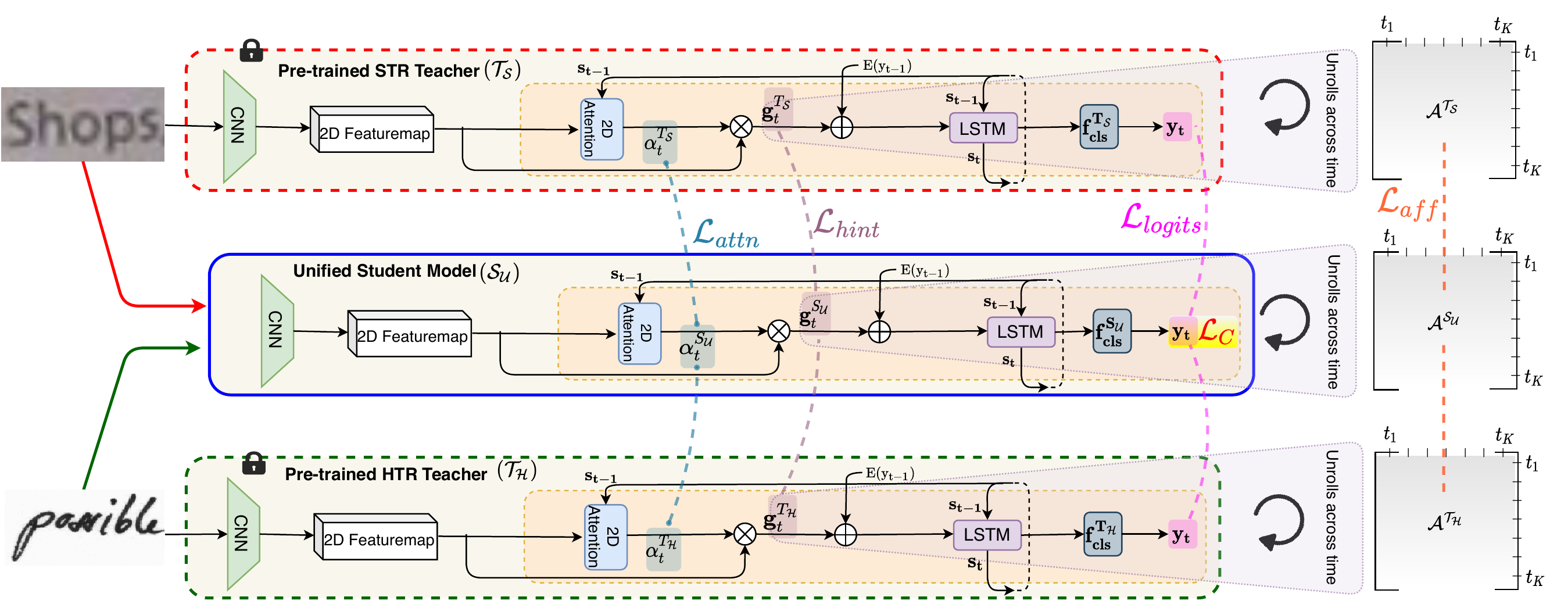}%{KD_1.pdf}
\end{center}
\vspace{-0.4cm}
    %\caption{STR and HTR models are first trained using only STR and HTR training images respectively via Cross-Entropy Loss ($\mathcal{L}_C$) (Left). Once trained, the STR and HTR Model are used as teacher models from which a unified model distils knowledge (Right) along with $\mathcal{L}_C$. }
    \caption{STR and HTR models, pre-trained using respective images, are used as \textit{teachers} to train a unified \textit{student} model via knowledge distillation, with four distillation losses and a cross-entropy loss ($\mathcal{L}_C$). The t$^{th}$ time-step for decoding is shown, which unrolls across time.}
%   STR and HTR models are first trained using only STR and HTR training images respectively via Cross-Entropy Loss ($\mathcal{L}_C$) (Left). Once trained, the STR and HTR Model are used as teacher models from which a unified model distils knowledge (Right) along with $\mathcal{L}_C$. This distillation process is enabled via four distillation loss: (a) Logits' Distillation loss ($\mathcal{L}_{logits}$) (b) Character Localised Hint Loss ($\mathcal{L}_{hint}$) (c) Attention Distillation Loss ($\mathcal{L}_{attn}$) and (d) Affinity Distillation Loss ($\mathcal{L}_{aff}$).
\vspace{-.2in}
\label{fig:Fig2}
\end{figure*}

\vspace{-0.1cm}
\section{Methodology}
\vspace{-0.15cm}
\noindent \textbf{Overview:} Our objective is to design a single unified model working both for STR (S) and HTR (H) word images. In this context, we have access to labelled STR datasets $\mathcal{D_S} = \{(\mathrm{I}_s, \mathrm{Y}_s) \in \mathcal{I}_s \times \mathcal{Y}_s\}$, as well as labelled HTR datasets $\mathcal{D_H} = \{(\mathrm{I}_h, \mathrm{Y}_h) \in \mathcal{I}_h \times \mathcal{Y}_h\}$. Here, $\mathrm{I}$ denotes word image from respective domain with label $\mathrm{Y} = \{y_1, y_2, \cdots, y_K \}$, and  $K$ denotes the variable length of ground-truth characters. We first train two individual text-recognition models using  $\mathcal{D_S}$ and  $\mathcal{D_H}$ independently. Thereafter, a single unified model is obtained from two domain specific teacher via knowledge distillation.

%  we describe generic text recognition model.. 
\vspace{-0.2cm}
\subsection{Baseline Text Recognition Model} \label{basemodel_textrecognition}
\vspace{-0.1cm}
Given an image $\mathrm{I}$, text recognition model $\mathcal{R}$ tries to predict the machine readable character sequence $\mathrm{Y}$.
% thus learning a conditional mapping $\mathcal{R}(\cdot): \mathrm{I} \rightarrow \mathrm{Y}$.  
Out of the two state-of-the-art choices dealing with irregular texts, we adopt 2-D attention that localises individual characters in a weakly supervised way, over complicated rectification network \cite{yang2019symmetry}. Our text recognition model consists of three components: (a) a backbone convolutional feature extractor \cite{shi2018aster}, (b) a RNN decoder predicting the characters autoregressively one at each time-step, (c) a 2D attentional block.

% Our text recognition model consists of three components: (a) a backbone convolutional feature extractor based on ResNet \cite{shi2018aster}, (b) a RNN decoder predicting the characters autoregressively one at a time step, (c) 2D attentional block that particularly focuses on specific spatial part of convolutional feature map that is relevant for predicting at a given time step. 

Let the extracted convolutional feature map be $\mathcal{F} \in \mathbb{R}^{h'\times w' \times d}$, where $h'$, $w'$ and $d$ signify height, width and number of channels. Every $d$ dimensional feature at $\mathcal{F}_{i,j}$ encodes a particular local image region based on the receptive fields. At every time step $t$, the decoder RNN predicts an output character or end-of-sequence (EOS) $y_{t}$ based on three factors: a) previous internal state $s_{t-1}$ of decoder RNN, (b) the character $y_{t-1}$ predicted in the last step, and (c) a glimpse vector $g_t$ representing the most relevant part of $\mathcal{F}$ for predicting $y_{t}$. To obtain $g_t$, previous hidden state $s_{t-1}$ acts as a query to discover the attentive regions as follows: 

\vspace{-0.6cm}
\begin{align} 
 & J =  \mathrm{tanh}(W_{F}\mathcal{F}_{i,j} + W_{\mathcal{B}} \circledast  \mathcal{F} + W_{s}s_{t-1}) \nonumber \\
& \alpha_{i,j}  =  \mathrm{softmax}(W_{a}^T  J_{i,j}) \label{eqn1} \\
& g_{t} =  \sum_{i,j} \alpha_{i,j} \cdot \mathcal{F}_{i,j}  \; \; i = [1, .., h'],  \; j = [1, .., w'] \label{eqn2}
\end{align}

\vspace{-0.2cm}
\noindent where, $W_{F}$, $W_{s}$, $W_{a}$ are the learnable weights. Calculating the attention weight $\alpha_{i,j}$ at every spatial position $(i,j)$, we employ a convolution operation ``$\circledast $" with $3\times3$ kernel $W_{\mathcal{B}}$ to consider the neighbourhood information in 2D attention mechanism. There exists $\alpha_t \in \mathbb{R}^{h' \times w'}$ corresponding to every time step of decoding, however  $t$ is dropped in Eqn. \ref{eqn1} and \ref{eqn2} for notational brevity.  
The current hidden state $S_{t}$ is updated by:  $\mathrm{\mathbf{(o_{t}, s_{t}) =  RNN(s_{t-1}; \; [E(y_{t-1}), \; g_{t}] ) )}}$, where $E(.)$ is character embedding layer with embedding dimension $\mathbb{R}^{128}$, and [.] signifies a concatenation operation. Finally, $\bm\tilde{y}_{t}$ is predicted as: $p(\bm\tilde{y}_{t})=\mathrm{softmax(W_{o}o_{t} + b_{o})}$ with learnable parameters $W_{o}$ and $b_{o}$.  This model is trained end-to-end using cross-entropy loss $\mathcal{H}(\cdot, \cdot)$ summed over the ground-truth sequence $Y = \{y_1, y_2, \cdots, y_K \}$, where $y_t$ is one-hot encoded vector of size $\mathrm{\mathbb{R}^{|V|}}$, and $|V|$ is the character vocabulary size. 
 \vspace{-0.3cm} 
\begin{equation}\label{eqn_ce}
\mathcal{L}_{C} = \sum_{t=1}^{K}\mathcal{H}(y_t, \bm\tilde{y_t})  = -\sum_{t=1}^{K} \sum_{i=1}^{|V|}y_{t,i} \log p(\bm\tilde{y}_{t,i})
 \vspace{-0.1cm} 
\end{equation}

\subsection{Basics: Knowledge Distillation} \label{basemodel_kd}
Initially, knowledge distillation (KD) \cite{hinton2015kd} was proposed for classification tasks to learn a  smaller student  model by mimicking the output of a pre-trained teacher. Given a particular data, let the output from pre-trained teacher be $\Tilde{y}^{T}_t = \mathrm{softmax}(l^T_t)$ and that of learnable student be $\Tilde{y}^{S}_t = \mathrm{softmax}(l^S_t)$,  where $l_t$ is pre-softmax logits from respective models. Temperature ($\tau$) normalised $\mathrm{softmax}$ is used to soften the output so that more information regarding inter-class similarity could be captured for training. Therefore, given  $\Tilde{y}^{T}_{\tau,t} = \mathrm{softmax}(\frac{l^T_t}{\tau})$, $\Tilde{y}^{S}_{\tau,t} = \mathrm{softmax}(\frac{l^S_t}{\tau})$ and ground-truth $y$, the student network is trained to optimise the following loss function: 

\vspace{-0.55cm}
\begin{equation}\label{basic_KD}
\mathcal{L}_{KD} = \sum_{t=1}^{K} \mathcal{H}(y_t,\Tilde{y}^{S}_t) + \lambda \sum_{t=1}^{K} \mathcal{H}(\Tilde{y}^{T}_{\tau,t} \ ,\Tilde{y}^{S}_{\tau,t})
 \vspace{-0.3cm}
\end{equation}

\noindent where $\lambda$ is a hyper-parameter balancing the two terms, and the first term signifies traditional cross-entropy loss between the output of student network and ground-truth labels, whereas the second term encourages the student to learn from softened output of teacher. 

Adopting basic KD formulation however is unsuitable for our purpose. Firstly, text recognition dealing with varied-length sequence recognition requires distilling local fine-grained character information. Additionally, there exists a sequential dependency among the predicted characters due to auto-regressive nature of attentional decoder, thus requiring a global consistency criteria during distillation process. (b) While training teacher and student usually involves same (single domain) dataset, we here have two separate domains, STR and HTR, which thus needs to deal with larger domain gap and data coming from two separate domains.

\subsection{Unifying Text Recognition Models}

\noindent \textbf{Overview:}  We propose a knowledge distillation method for sequential text images to unify both scene-text and handwriting recognition process into a \emph{single} model. Compared to traditional knowledge distillation, we have \emph{two} pre-trained teacher  networks ${T} \in \{{T}_\mathcal{S}, {T}_\mathcal{H} \}$, where ${T}_\mathcal{S}$ is a specialised model trained from \underline{$\mathcal{S}$}cene text images $\mathcal{D_S}$, and  ${{T}_\mathcal{H}}$ from \underline{$\mathcal{H}$}andwritten text images $\mathcal{D_H}$. Given these pretrained teachers, we aim to learn a single \underline{$\mathcal{U}$}nified  \underline{$\mathcal{S}$}tudent model ${S_{\mathcal{U}}}$ by \emph{four} distillation losses tailored for sequential recognition task, along with typical cross-entropy loss. ${{T}_\mathcal{S}}$, ${{T}_\mathcal{H}}$ and ${S_{\mathcal{U}}}$ all have identical architecture to text recognition network $\mathcal{R}(\cdot)$. Directly training a single model by including images from both the STR and HTR datasets leads to sub-optimal performance due to limited model capacity and large domain-gap. In contrast, training of \emph{specialised} models might assist to extract underlying structure from respective data, which can \emph{then} be distilled into a unified student network with guidance from the specialised teachers. 

We have two pre-trained teachers $\mathrm{{T} \in  \{ {T}_\mathcal{S}, {T}_\mathcal{H}}\}$, with images coming from two different  domains ${I} \in \{ {I_s}, {I_h}\}$. In order to train a student network ${S_{\mathcal{U}}}$, we will get one loss instance using STR pre-trained teacher and respective dataset  $({T}_\mathcal{S}, {I_s})$, and similarly  another loss term for HTR counterpart $({T}_\mathcal{H},{I_h})$. We describe the loss functions using generalised notation  $\mathrm{{(T,I)}}$ which basically has  two elements,  $({T}_\mathcal{S}, {I_s})$ and $({T}_\mathcal{H}, {I_h})$ respectively. Thus mathematically, $\mathrm{{(T,I)} : \{(T_\mathcal{S}, {I_s}), (T_\mathcal{H},{I_h}) \}}$. Please refer to Figure \ref{fig:Fig2}.
 
\keypoint{Logits\textquotesingle \space Distillation Loss:} We extend the traditional knowledge distillation loss for our sequence recognition task by aggregating cross-entropy loss over the sequence. Given an image ${I}$, let the temperature normalised softmax output from a particular pre-trained teacher and trainable student be  $\Tilde{y}_t^{{T}}({I})$ and  $\Tilde{y}_t^{{S}_{\mathcal{U}}}({I})$ at a particular time-step $t$. We ignore $\tau$ of Eqn.~\ref{basic_KD} here for notational brevity. We call this logits' distillation loss and define it as:  

\vspace{-0.3cm}
\begin{equation}
   \mathrm{\mathcal{L}_{logits}({T,I}) = \sum_{t=1}^{K} \mathcal{H}\left (  \Tilde{y}_t^{{T}}({I}), \; \Tilde{y}_t^{{S}_{\mathcal{U}}}({I}) \right  ) }
\vspace{-0.1 cm}
\end{equation}

\noindent where, $\mathrm{{(T,I)} : \{({T_\mathcal{S}}, {I_s}), ({T_\mathcal{H}},{I_h}) \}}$. We get two of such logits' distillation loss with respect to STR and HTR datasets (and pre-trained teachers) respectively. 

\keypoint{Character Localised Hint Loss:} The fact that intermediate features learned by the teacher could further act as a `hint' in the distillation process, was shown by Romero \etal \cite{romero2015fitnets}. Being a sequence recognition task however, text recognition needs to deal with variable length of sequence, with each character having variable width within itself.  While predicting every character, attention based decoder focuses on specific regions of convolutional feature-map. In order to circumvent the discrepancy due to variable character-width, we perform feature distillation loss at the space of character localised visual feature, termed as \emph{glimpse vector} (see Eqn. \ref{eqn2}) instead of global convolutional feature-map. This provides the teacher's supervision  at local level. As our student shares the same architecture identical to the pre-trained teachers, we do not need any parametric transformation layer to match the feature-space between them. The character localised hint loss is given by:   
\vspace{-0.3cm}
\begin{equation}
 \mathrm{\mathcal{L}_{hint}({T,I}) = \sum_{t=1}^{K}  \left \|   g_t^{{T}}({I})  -   g_t^{{S}_\mathcal{U}}({I}) \right \|_2 }
 \vspace{-0.15cm}
\end{equation}

\noindent where,  $\mathrm{{(T,I)} : \{({T_\mathcal{S}}, {I_s}), ({T_\mathcal{S}},{I_h}) \}}$. Given an input image $\mathrm{I}$, $g_t^{{T}}({I})$ and $g_t^{{S}_\mathcal{U}}({I})$ are glimpse vector of size $\mathbb{R}^d$ at $t$-th times step from a particular pre-trained teacher and trainable student.

\keypoint{Attention Distillation Loss:} While Character Localised Hint Loss aids in enriching the localised information (i.e. absolute information in the cropped region roughly enclosing the specific character), computed attention map (see Eqn \ref{eqn2}) brings \emph{contextual information} giving insights about which region is  \emph{relatively} more important than the others, over a convolutional feature map. Unlike attentional distillation, logits' distillation does not explicitly take into account the degree of influence each pixel has on model prediction, thus making the attention map computed at every step a complementary source of information \cite{zagoruyko2017KDattention} to learn from the student. Furthermore, HTR usually shows overlapping characters, which however rarely occurs in STR. Thus the student must learn the proper `look-back' (attention) mechanism from specialised teachers. Let $\alpha_t^{{T}}({I})$ and $\alpha_t^{{S}_\mathcal{U}}({I})$ represent the attention map from respective teacher and learnable student at $t$-th time step, both having size $\mathbb{R}^{h' \times w'}$ for  a given an input image $\mathrm{I}$. Considering  $\mathrm{{(T,I)} : \{({T_\mathcal{S}}, {I_s}), ({T_\mathcal{H}},{I_h}) \}}$, the attention distillation loss is computed as follows: 
\vspace{-0.3cm}
\begin{equation}
 \mathrm{\mathcal{L}_{attn}({T,I}) = \sum_{t=1}^{K}  \left \|   \alpha_t^{{T}}({I}) - \alpha_t^{{S}_\mathcal{U}}({I}) \right \|_2}
\vspace{-0.15cm}
\end{equation}

\noindent \textbf{Affinity Distillation Loss:} Attention based decoder encapsulates an implicit language model within itself, and the information of previously predicted characters flows through its hidden state. While previous character localised hint loss and attention distillation loss mostly contribute to information distillation at local level, with the later (attention) additionally contributing towards the contextual information, we need a global consistency loss to handle the long-range dependency among the characters. Thus we introduce an affinity distillation loss to model long-range non-local dependencies from the specialised teachers. Given character aligned features $\{g_{1}, g_{2}, \dots, g_{K}\}$ for a given image, the affinity matrix capturing the pair-wise correlation between every pair of characters is computed as: 
\vspace{-0.3cm}
\begin{equation}
      \mathcal{A}_{i,j} = \frac{1}{K \times K} \cdot \frac{g_{i}}{||g_{i}||_{2}} \cdot \frac{g_{j}}{||g_{j}||_{2}}
\vspace{-0.18cm}
\end{equation}
\noindent where, $\mathcal{A} \in \mathbb{R}^{K \times K}$  represents the affinity matrix for a word image having character sequence length $K$. We use $l_2$ loss to match the affinity matrix of specialised teacher 
$\mathcal{A}^{{T}}({I})$ and that of learnable student  $\mathcal{A}^{{S}_\mathcal{U}}({I})$: 
\vspace{-0.2cm}
 \begin{equation}
      \mathcal{L}_{aff}({T,I})  =  \left \|   \mathcal{A}^{{T}}({I})  -   \mathcal{A}^{{S}_\mathcal{U}}({I}) \right \|_2
\vspace{-0.18cm}
\end{equation}

\keypoint{Optimisation Procedure:}\label{sec-optimization} Apart from the four distillation loss in order to learn from the specialised teacher, the unified student model ${S}_\mathcal{U}$ is trained from ground-truth label for image $I \in \{I_s, I_h\}$ using typical cross-entropy loss (see Enq. \ref{eqn_ce}). Thus,  given $\mathrm{{(T,I)} : \{({T_\mathcal{S}}, {I_s}), ({T_\mathcal{H}},{I_h}) \}}$, the overall training objective for student becomes: 
\vspace{-0.3cm}
\begin{multline}\label{eq-all}
 \mathrm{\mathcal{L}_{all} = \sum_{\forall ({T, I})} \Big(\mathcal{L}_{C}({I})  + \lambda_1 \cdot  \mathcal{L}_{logits}({T,I})+  \lambda_2 \cdot \mathcal{L}_{attn}({T,I})} 
  \\[-10pt] \mathrm{+ \lambda_3 \cdot \mathcal{L}_{hint}({T,I}) + \lambda_4 \cdot \mathcal{L}_{aff}({T,I})\Big)}
 \vspace{-0.5cm}
\end{multline}

 \vspace{-0.3cm} 
Due to difference in complexity of the task of HTR and STR and their respective training data size, we observe a tendency to learn a biased model that over-fits on either STR or HTR dataset. To alleviate this, we employ a conditional distillation mechanism that stabilise training by deciding in what proportion to learn from two different individual specialised teacher that results in a unified student model performing ubiquitously over both STR and HTR scenarios.
% It is important to note that while supervision from a teacher enables superior learnability, a bad teacher is likely to have an adverse effect of limiting the performance. We therefore adopt the training mechanism of selective distillation \cite{tan2019multilingualKD} to aid the unified model in surpassing its teacher, as shown in Algorithm \ref{alg:train_loop}. The four distillation losses from a particular teacher are either removed when the student surpasses the teacher, or included once more if the performance of student worsens on further iterations. 
% \red{A detailed analysis on the signification of selective distillation is present in Appendix \ref{appendix:selectiveDistillation}}.
\setlength{\textfloatsep}{2pt}
\begin{algorithm}
    \caption{Training algorithm of the proposed framework}
    \label{alg:train_loop}
    \begin{algorithmic}[1]
        \State \textbf{Input:} Dataset: $\{\mathcal{D_S, D_H}\}$; Teachers: $\{ {T}_\mathcal{S}, {T}_\mathcal{H} \}$; Learning rate: $\eta$; Total Training Steps: $\mathcal{T}$, distil check: $\mathcal{T'}$; Accuracy metric: $\mathcal{A}cc$; distil acc. thresh. $\omega \geq 1$
        \State \textbf{Initialise:} Unified Student Model: $\mathcal{S_U}$, params: $\theta^{{S}_\mathcal{U}}$; Step: $t=1$; Gradient: $g$; Flags: $\{f^{\mathcal{S}}, f^{\mathcal{H}}\}$ are $True$
        \While{$t \leq \mathcal{T}$}
            \State $g=0$
            \State Get: $(\mathrm{I}_s, \mathrm{Y}_s) \in \mathcal{D_S}^{train}; (\mathrm{I}_h, \mathrm{Y}_h) \in \mathcal{D_H}^{train}$
            \State $g \mathrel{{+}{=}} \partial (\mathcal{L}_C(\mathrm{I}_s) + \mathcal{L}_C(\mathrm{I}_h)) / \partial \theta^{{S}_\mathcal{U}}$ \Comment{see eq. \ref{eqn_ce}}
            \For{each $\mathcal{L}_{KD}$ in $\mathcal{L}_{all}-\{\mathcal{L}_C\}$} \Comment{see eq. \ref{eq-all}}
                \State \algorithmicif{$f^{\mathcal{S}}$} \algorithmicthen\ $g \mathrel{{+}{=}} \partial \mathcal{L}_{KD}({T}_\mathcal{S}, \mathrm{I}_s) / \partial \theta^{{S}_\mathcal{U}}$
                \State \algorithmicif{$f^{\mathcal{H}}$} \algorithmicthen\ $g \mathrel{{+}{=}} \partial \mathcal{L}_{KD}({T}_\mathcal{H}, \mathrm{I}_h) / \partial \theta^{{S}_\mathcal{U}}$
            \EndFor
            \State Update $\theta^{{S}_\mathcal{U}}: \ \theta^{{S}_\mathcal{U}} = \theta^{{S}_\mathcal{U}} - \eta * g$
            \If{$t \% \mathcal{T'} == 0$} \Comment{\emph{conditional distillation}}
                \State $\mathcal{L} = \mathcal{L}_{all} - \{\mathcal{L}_{C}\}$
                \State $\{\mathcal{I}_s^{val}, \mathcal{Y}_s^{val}\} = \mathcal{D_S}^{val}$; $\{\mathcal{I}_h^{val}, \mathcal{Y}_h^{val}\} = \mathcal{D_H}^{val}$
                \State \algorithmicif{$\mathcal{L}(T_{\mathcal{S}},\mathcal{I}_s) > \omega \cdot \mathcal{L}(T_{\mathcal{H}},\mathcal{I}_h)$} \algorithmicthen\ $f^{\mathcal{H}}=False$ 
                \State \algorithmicelse\ $f^{\mathcal{H}}=True$
                \State \algorithmicif{$\mathcal{L}(T_{\mathcal{H}},\mathcal{I}_h) > \omega \cdot \mathcal{L}(T_{\mathcal{S}},\mathcal{I}_s)$} \algorithmicthen\ $f^{\mathcal{S}}=False$ 
                \State \algorithmicelse\ $f^{\mathcal{S}}=True$
                % \State \algorithmicif{$\mathcal{A}cc(\mathcal{S_U}) < \mathcal{A}cc({T}_\mathcal{S}) + \omega$} \algorithmicthen\ $f^{\mathcal{S}}=True$ 
                % \State \algorithmicelse\ $f^{\mathcal{S}}=False$
                % \State \algorithmicif{$\mathcal{A}cc(\mathcal{S_U}) < \mathcal{A}cc({T}_\mathcal{H}) + \omega$} \algorithmicthen\ $f^{\mathcal{H}}=True$ 
                % \State \algorithmicelse\ $f^{\mathcal{H}}=False$
            \EndIf
            \State $t=t+1$
        \EndWhile
    \end{algorithmic}
\end{algorithm}

\vspace{-0.2cm}
\section{Experiments}\label{sec:experiments}
%\subsection{Experimental Setup}\label{setup}
\vspace{-0.15cm}
\noindent \textbf{Datasets:} 
Training paradigm for STR involves using large synthetic datasets such as Synth90k \cite{jaderberg2014synthetic} and SynthText \cite{SynthText} with $8$ and $6$ million images respectively, and evaluating (without fine-tuning) on real images such as: \noindent{\textbf{IIIT5K-Words}}, \noindent{\textbf{Street View Text (SVT)}}, \noindent{\textbf{SVT-Perspective (SVT-P)}}, \noindent{\textbf{ICDAR 2013 (IC13)}}, \noindent{\textbf{ICDAR 2015 (IC15)}}, and \noindent{\textbf{CUTE80}}. IIIT5-K Words \cite{IIIT5K-Words} has 5000 cropped words from Google image search. SVT \cite{WangICCV2011} hosts $647$ images collected from Google Street View where most images are blurry, noisy and have low resolution. SVT-P \cite{SVT-P} has $639$ word images also taken from Google Street view but with side-view snapshots resulting in severe perspective distortions. ICD13 \cite{ICDAR2013} contains $848$ cropped word patches with mostly regular images unlike IC15 \cite{ICDAR2015} which has 2077 word images that are irregular i.e. oriented, perspective or curved. Unlike others, CUTE80 \cite{CUTE80} dataset contains high resolution image but have curved text. In context of HTR, we follow the evaluation setup described in \cite{bhunia2019handwriting} on two large standard datasets viz, \noindent{\textbf{IAM}} \cite{IamDataset} (1,15,320 words) and \noindent{\textbf{RIMES}} (66,982 words).

% we consider IAM \cite{IamDataset} dataset consisting of $1,15,320$ handwritten words which allows for both training and testing.

\noindent \textbf{Implementation Details:} We use a 31-layer CNN backbone feature extractor \cite{2dAtten2019aaai} without any pre-training. The input image is resized to $48\times160$ following \cite{2dAtten2019aaai}. We first pre-train the specialised HTR and STR model at a time. For STR, we use Synth90k \cite{jaderberg2014synthetic} and SynthText \cite{SynthText} dataset together, and respective training set is used for experiments on IAM and RIMES dataset individually. We use Adam optimiser with initial learning rate of $0.001$ and batch size of 32 for both specialised teacher pre-training, and distillation based unified student model training. Decay rate of $0.9$ is applied after every $10^{4}$ iteration till the learning rate drops to $10^{-5}$. During conditional distillation (Algorithm \ref{alg:train_loop}), loss is compared over the validation set with $\omega=1.05$. \doublecheck{We set $\lambda_1$, $\lambda_2$, $\lambda_3$, and $\lambda_4$ as $0.5$, $5$, $1$ and $1$  respectively.} We implement the network and its training paradigm using PyTorch trained in a 11 GB NVIDIA RTX-2080-Ti GPU.

\noindent \textbf{Evaluation Protocol:}
To better understand the challenges of unifying STR and HTR, and recognise contribution of each alternative training paradigm we evaluate as follows: (i) we first evaluate the pre-trained teacher models on the dataset for what it has been trained for, e.g. $\mathcal{T_S}$ on testing set of STR dataset, and $\mathcal{T_H}$ on that of HTR dataset. (ii) Next, we evaluate on the alternative dataset for pre-trained teacher model and see how the performance drops in cross-dataset scenarios, e.g.  $\mathcal{T_S}$ on testing set of HTR dataset, and vice-versa. ii) Finally, we evaluate the unified student model $\mathcal{S_U}$ on both STR and HTR datasets to verify if a single model can perform ubiquitously for both scenarios.  

\setlength{\tabcolsep}{2.0pt}
\begin{table*}[hbt!]
    \begin{minipage}{0.55\linewidth}
        % \centering
        \caption{Quantitative performance against various alternatives. \textbf{Competitors use \emph{combined} STR+HTR datasets} in different setups: (a) Multi-Task (Joint) Training, (b) Unsupervised and Supervised Domain Adaptation (DA), (c) Domain Generalization (DG). }
        \scriptsize
        \begin{tabular}{c|c|c|c|c|c|c|c|c}
            \hline
            \multirow{2}{*}{\textbf{Methods}} & \multicolumn{6}{c|}{\textbf{STR datasets}} & \multicolumn{2}{c}{\textbf{HTR dataset}} \\
            \cline{2-9}
            & IIIT5-K & SVT & IC13 & IC15 & SVT-P & CUTE80 & IAM & RIMES \\
            \hline
            Multi-Task-Training-(I)  & 86.1 & 83.6 & 87.2 & 70.4 & 77.8 & 79.4 & 81.8 & 86.2 \\
            Multi-Task-Training-(II)  & 35.4 & 34.5 & 36.3 & 29.1 & 32.1 & 32.5 & 81.9 & 85.9 \\
            Multi-Task-Training-(III)  & 83.2 & 80.5 & 84.1 & 67.1 & 74.1 & 76.3 & 77.9 & 82.3 \\
            DA-Adv-Unsup (STR $\rightarrow$ HTR) & 82.6 & 80.1 & 84.2 & 66.8 & 74.2 & 75.8 & 58.7 & 64.1 \\
            DA-Adv-Unsup (HTR $\rightarrow$ STR) & 16.6 & 12.9 & 15.4 & 12.1 & 12.7 & 13.4 & 78.1 & 82.4 \\
            % DA-Adv-Unsup (HTR $\rightarrow$ STR) & 87.3 & 84.9 & 88.4 & 71.7 & 79.1 & 80.8 & 82.6 & 87.0 \\
            DA-Adv-Sup & 88.1 & 85.6 & 89.2 & 72.5 & 79.9 & 81.6 & 83.1 & 87.5 \\
            DA-Corr-Unsup (STR $\rightarrow$ HTR) & 82.7 & 80.2 & 84.5 & 67.8 & 74.7 & 76.1 & 82.7 & 87.1 \\
            DA-Corr-Unsup (HTR $\rightarrow$ STR) & 17.1 & 13.1 & 15.9 & 12.7 & 13.1 & 13.9 & 82.7 & 87.1 \\
            % DA-Corr-Unsup (HTR $\rightarrow$ STR) & 87.4 & 84.9 & 88.5 & 71.8 & 79.2 & 80.9 & 82.7 & 87.1 \\
            DA-Corr-Sup & 88.3 & 85.8 & 89.4 & 72.7 & 80.1 & 81.8 & 83.2 & 87.6 \\
            DG-training & 88.5 & 86.0 & 89.5 & 72.9 & 80.3 & 82.0 & 83.4 & 87.7 \\
            \hline
            \textbf{Proposed} & 92.3 & 89.9 & 93.3 & 76.9 & 84.4 & 86.3 & 86.4 & 90.6 \\
            \hline
        \end{tabular}
        \label{tab:competitors}
    \end{minipage}
    \hfill
    \begin{minipage}{0.40\linewidth}
        \centering
        \caption{Quantitative comparison of our STR-\textit{only} and HTR-\textit{only} models, trained on STR and HTR datasets respectively, against state-of-the-arts. Our method uses STR-only and HTR-only as teachers during KD.}
        \scriptsize
        \begin{tabular}{c|c|c|c|c|c|c}
            \hline
            \multirow{2}{*}{\textbf{Methods}} & \multicolumn{4}{c|}{\textbf{STR datasets}} & \multicolumn{2}{c}{\textbf{HTR dataset}} \\
            \cline{2-7}
             & IIIT5-K & SVT & IC13 & IC15 & IAM & RIMES \\
            \hline
            Shi \etal \cite{shi2018aster} & 93.4 & 93.6 & 91.8 & 76.1 & -- & -- \\
            Baek \etal \cite{baek2019wrong} & 87.9 & 87.5 & 92.3 & 71.8 & -- & -- \\
            % Li \etal \cite{2dAtten2019aaai} & 95.0 & 91.2 & 94.0 & 78.8 & -- & -- \\
            Yu \etal \cite{yu2020towards} & 94.8 & 91.5 & 95.5 & 82.7 & -- & -- \\
            Litman \etal \cite{litman2020scatter} & 93.7 & 92.7 & 93.9 & 82.2 & -- & -- \\
            Bhunia \etal \cite{bhunia2019handwriting} & -- & -- & -- & -- & 82.81 & 88.53 \\
             \hline
            STR-only Model & 93.1 & 90.9 & 93.5 & 78.2 & 53.4 & 58.5 \\
            HTR-only Model & 11.5 & 7.6 & 10.3 & 7.1 & 85.9 & 90.2 \\
            \doublecheck{Joint STR-HTR Model} & 86.1 & 83.6 & 87.2 & 70.4 & 81.8 & 86.2 \\
             \hline
            \textbf{Proposed (Unified)} & 92.3 & 89.9 & 93.3 & 76.9 & 86.4 & 90.6 \\
             \hline
        \end{tabular}
        \label{tab:comparison}
	\end{minipage}
\vspace{-0.5cm}
\end{table*}

\vspace{-0.1cm}
\subsection{Competitors}
\vspace{-0.1cm}
To the best of our knowledge, there has been no prior work dealing with the objective of unifying STR and HTR models into a single model. Thus, we design a few strong baselines based on the existing literature by our own. \emph{(i)} \textbf{Multi-Task-Training:} This is a naive \emph{frustratingly easy} training paradigm \cite{daume2009frustratingly} where samples belonging to both STR and HTR datasets are used to train a single network guided by cross-entropy loss. \doublecheck{Since STR has overwhelmingly large synthetic training samples \cite{jaderberg2014synthetic, SynthText} compared to HTR dataset \cite{IamDataset}, we use weighted random sampling (variant-I) to balance training data. Conversely, we randomly sample a subset from STR dataset (variant-II) to forcefully make the number of training images similar for HTR and STR datasets in order to validate the utility of conditional distillation. In variant-III, we treat HTR and STR character units as different classes, thus extending it to N-class to 2N class classification at each time step.} \textbf{\emph{(ii)} DA-Corr-Unsup:} An obvious alternative is to try out any domain adaptation method introduced for sequence recognition task. Zhang \etal \cite{zhang2019cvprDAtext} proposed unsupervised domain adaptation (DA) technique for text images.
 We start by training a model on either STR (or HTR) images that acts as our source domain, followed by unsupervised adaptation to the target HTR (or STR) images -- thus we have two version of this model STR model adapted to HTR as (HTR $\mapsto$STR), and (STR $\mapsto$HTR). 
 %  \doublecheck{We train the unified network by alternatively considering either STR (or HTR) images as the labelled source domain while the other, HTR (or STR), acts as the unlabelled target.} 
 Second-order statistics-correlation distance \cite{sun2016deep} is used to align feature distribution from two domain. 
%  Along with cross-entropy loss for the source domain, it aims to align the distributions of the source and target domain at a localised character level feature space using CORAL \cite{sun2016deep} that computes second-order statistics-correlation distance.
 %  Distribution matching is performed over glimpse vectors using CORAL \cite{sun2016deep} that computes second-order statistics-correlation distance between source and target domain.
 \textbf{[\emph{iii}] DA-Corr-Sup:} As we have the access to both labelled STR and HTR datasets, we further extend the unsupervised DA setup of  Zhang \etal \cite{zhang2019cvprDAtext} by considering target domain to be annotated, allowing supervised DA.
Cross-entropy loss is minimised for both source and target domain in association to second-order statistics-correlation between both STR and HTR domains. \textbf{[\emph{iv}] DA-Adv-Unsup:} We further adopt a recent work by Kang \etal \cite{kang2020DApool} employing adversarial learning for unsupervised domain adaptation for text recognition. Here, the setup remains same as  DA-Corr-Unsup having two versions as (HTR $\mapsto$STR) and (STR $\mapsto$HTR), but domain adaptation tackled through a discriminator with a preceding gradient-reversal layer.  \textbf{\emph{(v)} DA-Adv-Sup:} This is again a similar adaptation of \cite{kang2020DApool} following supervised DA which minimise Cross-Entropy and domain classification loss for both STR and HTR.  \textbf{\emph{(vi)} DG-Training:} Another alternative way to address this problem could be to use Domain Generalisation (DG) training based on model agnostic meta-learning using episodic-training \cite{finn2017MAML}. It involves using weighted ($\lambda$) summation \cite{guo2020learning} for gradient (over meta-train set) and meta-gradient (over meta-test split through inner loop update) to train our baseline text recognition model. The inner-loop update process consists of support set consisting images of either STR (or HTR) word images while the outer-loop update process is materialised using images from a different domain i.e. HTR (or STR). Such inner and outer-loop based optimisation strategy helps learn a model that aims to generalise well for both scenarios without further fine-tuning. 

\begin{figure}[t!]
\begin{center}
  \includegraphics[width=\linewidth]{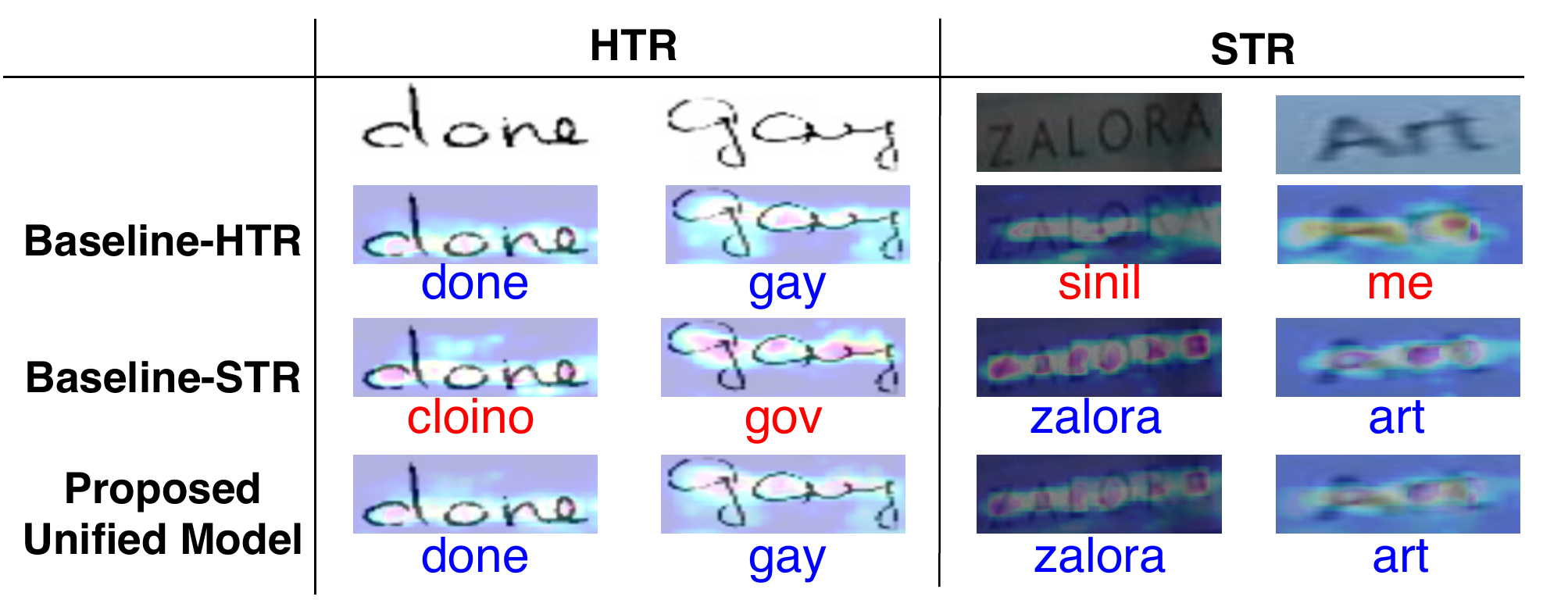}
\end{center}
\vspace{-.50cm}
%   \caption{Probability of occurrence of $30$ common words in IC15 \cite{ICDAR2015} and IAM \cite{IamDataset} datasets. Note, while some words such as \emph{``shops"}, frequently occurs in scene dataset \cite{ICDAR2015}, are not so frequent in handwritten dataset \cite{IamDataset}. This provides an empirical evidence of language diversity between STR and HTR datasets.}
% \caption{Word cloud of STR (including all STR datasets) and HTR (IAM) datasets show the vocabulary diversity.}
\caption{Illustrative examples with attention maps, and prediction ($\mathrm{\textcolor{red}{red} \rightarrow incorrect}$, $\mathrm{\textcolor{blue}{blue} \rightarrow correct}$). While discrepancy exists for cross-dataset scenarios, attention-map from unified model is nearly consistent with that of respective specialised model.}
\vspace{.10cm}
\label{fig:atten_distill}
\end{figure}

\vspace{-0.1cm}
\subsection{Performance Analysis}
 \vspace{-0.1cm}
% 721
% From Table \ref{tab:competitors} and \ref{tab:comparison}, it can be seen that our KD based unifying approach for multi-scenario text recognition outperforms all other baselines by a significant margin. 
From Table \ref{tab:comparison}, it can be seen that while a model trained on HTR fails miserably when evaluated on STR datasets, training on STR followed by testing on HTR does not result in a similar collapse in performance. This indicates that although STR scenarios partially encompass domain specific HTR attributes, the reverse is not true. Interestingly, this is likely why there is a positive transfer for HTR datasets using \emph{unified model} compared to HTR-only counterpart. Moreover, our KD based unifying approach for multi-scenario text recognition outperforms all other baselines by a significant margin.
In particular, \textbf{(i)} For baselines designed for unification, we attribute the limitation of all three multitask-learning-training (also named as joint-training) variants to the reason that it does not consider the varying complexity of two different tasks during joint training. Instead, our pre-trained teacher models first discover the \emph{specialised knowledge} from respective scenario. Given the specialised knowledge, our framework can encapsulate it into a single framework by balancing the learning via \emph{conditional distillation} from two different data sources (see Figure \ref{fig:atten_distill}). We outperform this joint-training (variant-I being the best performing competitor) baseline by a margin of almost $6-7\%$ on every dataset. Limited performance of variant-II validates the necessity and motivation of conditional distillation. \textbf{(ii)} The performance of unsupervised DA is limited by a significant margin while evaluating on both HTR and STR datasets. Starting from any source domain, it hardly gives any significant rise in target domain, rather the performance even decreases in the source domain after adaptation. An inevitable corollary of unsupervised DA is the lack of any guarantee that a model will retain information about source domain after successful adaptation to the target domain. \textbf{(iii)} The Domain Adaptation (DA) based pipelines suppress multitask-learning-training baseline while using supervised-labels from both the datasets, but lags behind us by $3.5-4.5\%$ on an average. Even using supervised-labels from both the datasets, the learning process oscillates around discovering domain invariant representation, and ignores main objective of unification of two specialised knowledge available from labelled datasets. Furthermore, adversarial learning based DA \cite{kang2020DApool} falls short compared to covariance based character-wise distribution alignment \cite{zhang2019cvprDAtext} for text recognition -- this also supports our design of using distillation loss over glimpse vectors.   \textbf{(iv)} Both \cite{zhang2019cvprDAtext} and \cite{wan2020vocabularyReliance} train a text recognition model on a source domain comprising of easily available synthetic images followed by unsupervised adaptation to target domain consisting of real world text images. While cost-effective training from synthetic-data is their major objective, we consider to have access to both the labelled datasets (which are readily available nowadays) to design an unified model working for both scenarios -- making our work orthogonal to these two DA based pipelines.  \textbf{(v)}  The purpose of Domain Generalisation (DG) is to find a model robust to domain-shift, giving satisfactory performance without the need of further adaptation. While such technique play a key role in unseen data regime, given enough labelled data, a frustratingly-simpler \cite{daume2009frustratingly} alternative -- multi-task learning -- also achieves similar performance gains. Given the labelled STR and HTR training data, we observe that although DG-training outperforms multi-task-training, it lags behind our proposed method by almost $4\%$ due to unavailability of privilege information (Table \ref{tab:competitors}). 
\textbf{(vi)} \cut{Furthermore, we attribute one more challenge behind the multi-scenario text unification approach -- the vocabulary diversity between STR and HTR scenarios. Importance of vocabulary (words present in the dataset) for state-of-the-art text recognition methods is apparent by their poor performance to images with words outside vocabulary \cite{wan2020vocabularyReliance}. We have observed that nouns such as `stop', `walk', `bar' etc are mostly found on placards and road signs (hence STR) while verbs or adverbs like `taking', `giving' or `nicely' appear more in HTR. Due to this discrepancy, our specialised knowledge discovery is important which is followed by KD based unification.} The diversity of vocabulary (words present in the dataset) between STR and HTR scenarios forms an important limitation to achieve SOTA performance \cite{wan2020vocabularyReliance}. While nouns (`stop', `walk') are observed in STR images (placard, road signs), verbs or adverbs (`taking', `giving') are more prevalent in HTR. Our specialised knowledge discovery bridges this discrepancy via unification.

\vspace{-0.2cm}
\setlength{\tabcolsep}{1pt}
\begin{table}[hbt!]
    \begin{minipage}{0.50\linewidth}
    \centering
     \scriptsize
    \caption{ {Contribution (WRA) of each KD constraint with $\mathcal{L}_{C}$}}
    % given by: (i) $\mathcal{L}_{logits}$, (ii) $\mathcal{L}_{attn}$, (iii) $\mathcal{L}_{hint}$, (iv) $\mathcal{L}_{aff}$, (v) using all constraints.
    \begin{tabular}{cccccc}
        \hline
        $\mathcal{L}_{logits}$ & $\mathcal{L}_{attn}$ & $\mathcal{L}_{hint}$ & $\mathcal{L}_{aff}$ & IC15 & IAM \\ \hline
        - & - & - & - & 70.4 & 81.8 \\
        \checkmark & - & - & - & 75.3 & 84.9 \\
        \checkmark & \checkmark & - & - & 75.7 & 85.3 \\
        \checkmark & \checkmark & \checkmark & - & 76.4 & 85.9 \\ \hline
        % \checkmark & \checkmark & \checkmark & \checkmark & 75.2 & 84.8 \\ \hline
        \checkmark & \checkmark & \checkmark & \checkmark & 76.9 & 86.4 \\
        \hline
    \end{tabular}
    \label{tab:ablation}
    \end{minipage}
    \hfill
    \begin{minipage}{0.44\linewidth}
    \centering
    \caption{Analysis of Time and Space complexities. \cut{between alternative design choices.}}
    \scriptsize
    \begin{tabular}{ccccc}
        \hline
        Methods & IC15 & IAM & GFlops & Params. \\
        \hline
        M.T.T & 70.4 & 81.8 & 0.67 & 19M \\
        B.C.R & 74.4 & 83.1 & 0.80 & 50M \\
        KD-Res-12 & 74.2 & 83.9 & 0.38 & 16M \\
        KD-Res-31 & 74.7 & 84.2 & 0.12 & 9M \\
        \hline
        \textbf{Proposed} & 76.9 & 86.4 & 0.67 & 19M \\
        \hline
    \end{tabular}
    \label{tab:computational}
    \end{minipage}
\vspace{-0.2cm}
\end{table}

\vspace{-0.15cm}
\subsection{Ablation Study:} 
\vspace{-0.15cm}
\noindent \textbf{[i] Competitiveness of our baseline:} Our baseline text recognition model is loosely inspired from the work by Li \etal \cite{2dAtten2019aaai}  that also uses 2D attention to locate the characters in weakly supervised manner even from irregular text images for recognition. An alternative is to use a two-stage framework consisting of an \emph{image rectification module} \cite{shi2018aster} followed by text recognition \cite{baek2019wrong}. But as observed by Zhang \etal \cite{zhang2019cvprDAtext}, although rectification based networks designed to handle spatial distortions lead to good performance in irregular STR datasets, it becomes a bottleneck for HTR tasks due to distortion caused by handwriting styles.  Hence, for the purpose of unified text recognition, 2D attention mechanism provides a reasonable choice to bypass the rectification network in the text recognition system. Table \ref{tab:comparison} shows our baseline text recognition model to have a competitive performance in comparison to existing methods in both STR and HTR datasets. Moreover, we tried to replicate our KD based pipeline incorporating \emph{image rectification module} on the top of \cite{baek2019wrong}, but performance gets limited to $75.9\%$ and $85.5\%$ on IC15 and IAM dataset, respectively. \textbf{[ii] Binary-Classifier based two-stage alternative:} Besides \emph{Multi-Task-Training} (\textbf{M.T.T}), another alternative is to use a binary-classifier (\textbf{B.C.R}) to classify between HTR and STR samples, then followed by selecting either STR or HTR model accordingly. While this achieves comparable performance with ours, it involves heavy computational expenses for maintaining three networks ($2$ specialised models + 1 classifier) together even while using simple ResNet18 as binary classifier -- thus making it inefficient for online deployment. A thorough analysis on the computational aspect is shown in Table \ref{tab:computational}.
\textbf{[iii] Significance of individual losses:} Among the four knowledge distillation losses ($\mathrm{\mathcal{L}_{logits}, \mathcal{L}_{attn}, \mathcal{L}_{hint}, \mathcal{L}_{aff}}$), we use one of these distillation constraints along with $\mathrm{\mathcal{L_C}}$ to understand their individual relative contribution. Table \ref{tab:ablation} shows $\mathcal{L}_{hint}$ to have the greatest impact among others, increasing accuracy on IC15 (IAM) by $5.1\%$  $(3.3\%)$, followed by $\mathrm{\mathcal{L}_{logits}}$ resulting in an increase of $4.9\%$ $(3.1\%)$, $\mathrm{\mathcal{L}_{aff}}$ by $4.8\%$ $(3.0\%)$ and $\mathrm{\mathcal{L}_{attn}}$ by $4.3\%$ $(2.6\%)$.   \textbf{[iv] Significance of conditional distillation:} Besides the wide difference in training data size, the complexity of the task of HTR and STR is different. A simple multi-task-training often over-fits on either STR or HTR dataset -- leading to sub-optimal performance of the unified student model. Thus, conditional distillation not only stabilises training, but also helps the student model to decide in what proportion to learn from two different individual specialised teachers, so that the unified model performs ubiquitously over both STR and HTR scenarios. Without conditional distillation, the performance is reduced by $2.5\%$ and $0.4\%$ on IC15 and IAM datasets, respectively. \doublecheck{The hyperparameter $\omega$ controlling the conditional distillation process is varied at $1.01, 1.03, 1.05, 1.07, 1.10$, and results on IC15 (IAM) are $76.8\%$ $(86.3\%)$, $76.9\%$ $(86.3\%)$, $76.9\%$ $(86.4\%)$, $76.8\%$ $(86.4\%)$, $76.8\%$ $(86.4\%)$}. 
% The hyperparameter $\varepsilon$ controlling the conditional distillation process is varied and evaluated in Table \ref{tab:computational}.  
\textbf{[vi] Hint Loss location:} While hint-based training leads to performance enhancements, the location of feature distillation loss is debatable based on the model's architecture. Thus, we employ $\mathcal{L}_{hint}$ on: (a) CNN features $\mathcal{F}$  and (b) hidden state $s_t$ of attentional decoder. \cut{table. We employ $\mathcal{L}_{hint}$ on: (a) CNN features $\mathcal{F}$ (b) RNN hidden features $S_t$.} Using $\mathcal{L}_{hint}$ on $\mathcal{F}$ lead to a performance improvement of $3.8\%$ ($2.2\%$) while on $s_t$ results in $4.6\%$ ($2.5\%$) enhancement on IC15(IAM) datasets; both of which are lower as compared to $\mathcal{L}_{hint}$ on context vector $g$ giving $5.1\%$ ($3.3\%$) improvement over the baseline model.  \textbf{[vii] Reduce model size using KD:} Knowledge distillation is a generic  method used to compress \cite{hinton2015kd} any deep model regardless of the structural difference between teacher and student. Hence, we further check if our tailored KD method for attentional decoder based text recognition framework could be used off-the-shelf to reduce the model size of unified student. We replace our student model having 31-layer ResNet with just 12-layer (2+2+3+3+2) as KD-ResNet-12, and replace normal convolution by depth-wise convolution following MobileNetV2 architecture \cite{sandler2018mobilenetv2} to obtain  KD-ResNet-31. The two resulting light-weight architectures give $74.2\%$  $(83.9\%)$ and $74.7\%$ $(84.2\%)$ accuracies in IC15 (IAM) datasets without much significant drop compared to our full version as shown in Table \ref{tab:computational}. This suggests that our framework could be widened further for model compression of text recognition model. 

\vspace{-0.3cm} 

\section{Conclusion}
\vspace{-0.1cm}
We put forth a novel perspective towards text recognition -- unifying multi-scenario text recognition models. To this end we introduced a robust resource-economic online serving solution by proposing a knowledge distillation based framework employing four distillation losses to tackle the varying length of sequential text images. This helps us reduce the domain gap between scene and handwritten images while alleviating language diversity and model capacity limitations. The resulting unified model proves capable of handling both scenarios, performing at par with individual models, even surpassing them at times (e.g. in HTR).
% comparable and even slightly better results than individual models.

{\small
\bibliographystyle{ieee_fullname}
\bibliography{egbib}
}

\end{document}